%% file: arxiv_melee.tex
\documentclass{article}

\usepackage{microtype}
\usepackage{graphicx}
\usepackage{subfigure}
\usepackage{booktabs} %
\usepackage{xspace}
\usepackage{haldefs}
\usepackage[inline]{enumitem}
\usepackage{dsfont}
\usepackage{wrapfig}

\usepackage{hyperref}

\usepackage[accepted]{icml2019}

\icmltitlerunning{Meta-Learning for Contextual Bandit Exploration}

\newcommand{\ourname}{\textsc{M\^el\'ee}\xspace}
\newcommand{\ournamelong}{MEta LEarner for Exploration\xspace}

\newcommand{\polopt}{\textsc{PolOpt}\xspace}
\newcommand{\reg}{{\text{Reg}}}

\newcommand{\gstar}{{\g^\star}}

\newcommand{\g}{\pi}

\def\equationautorefname~#1\null{Eq~#1\null}
\renewcommand{\sectionautorefname}{\S\kern-0.2em}
\renewcommand{\subsectionautorefname}{\S\kern-0.2em}
\renewcommand{\subsubsectionautorefname}{\S\kern-0.2em}
\makeatletter \newcommand{\ALC@uniqueautorefname}{line} \makeatother

\newcommand{\algorithmname}[1]{\textsc{#1}\xspace}
\newenvironment{proof}{\paragraph{Proof:}}{\hfill$\square$}

\newcommand{\1}{\mathds{1}}
\newcommand{\R}{\mathbb{R}}

\newtheorem{theorem}{Theorem}

\begin{document}

\twocolumn[
\icmltitle{Meta-Learning for Contextual Bandit Exploration}

\icmlsetsymbol{equal}{*}

\begin{icmlauthorlist}
\icmlauthor{Amr Sharaf}{umd}
\icmlauthor{Hal Daum\'e III}{umd,msr}
\end{icmlauthorlist}

\icmlaffiliation{umd}{Department of Computer Science, University of Maryland, College Park, Maryland, USA}
\icmlaffiliation{msr}{Microsoft Research, New York City}

\icmlcorrespondingauthor{Amr Sharaf}{amr@cs.umd.edu}

\icmlkeywords{Machine Learning, ICML}

\vskip 0.3in
]

\printAffiliationsAndNotice{}  %

\begin{abstract}

\input{abstract}

\end{abstract}

\section{Introduction} \label{sec:intro} \input{intro}

\section{Meta-Learning for Contextual Bandits} \label{sec:approach} \input{approach}

\section{Theoretical Guarantees} \label{sec:guarantees} \input{guarantees}

\section{Experimental Setup and Results} \label{sec:experiments} \input{experiments}

\section{Related Work and Discussion} \label{sec:related} \input{related}

\bibliography{references}
\bibliographystyle{icml2019}

\clearpage
\appendix
\begin{centering}
  \Large
  Supplementary Material For:\\
  Meta-Learning for Contextual Bandit Exploration\\
\end{centering}

\makeappendix

\end{document}

%% file: abstract.tex
We describe \ourname, a meta-learning algorithm for learning a good exploration policy in the interactive contextual bandit setting.
Here, an algorithm must take actions based on contexts, and learn based only on a reward signal from the action taken, thereby generating an exploration/exploitation trade-off.
\ourname addresses this trade-off by learning a good exploration strategy for offline tasks based on synthetic data, on which it can simulate the contextual bandit setting.
Based on these simulations, \ourname uses an imitation learning strategy to learn a good exploration policy that can then be applied to true contextual bandit tasks at test time.
We compare \ourname to seven strong baseline contextual bandit algorithms on a set of three hundred real-world datasets, on which it outperforms alternatives in most settings, especially when differences in rewards are large.
Finally, we demonstrate the importance of having a rich feature representation for learning how to explore.

%% file: intro.tex
%

In a contextual bandit problem, an agent attempts to optimize its behavior over a sequence of rounds based on limited feedback~\citep{kaelbling1994associative,auer2002ucb,langford2008epoch}.
In each round, the agent chooses an action based on a context (features) for that round, and observes a reward for that action but no others (\autoref{sec:cb}).
Contextual bandit problems arise in many real-world settings like online recommendations and personalized medicine.
As in reinforcement learning, the agent must learn to balance \emph{exploitation} (taking actions that, based on past experience, it believes will lead to high instantaneous reward) and \emph{exploration} (trying actions that it knows less about).

In this paper, we present a meta-learning approach to automatically learn a good exploration mechanism from data. %
To achieve this, we use synthetic supervised learning data sets on which we can simulate contextual bandit tasks in an offline setting.
Based on these simulations, our algorithm, \ourname (\ournamelong)\footnote{\textbf{Code release:} the code is available online \url{https://www.dropbox.com/sh/dc3v8po5cbu8zaw/AACu1f_4c4wIZxD1e7W0KVZ0a?dl=0}}, learns a good heuristic exploration strategy that should ideally generalize to future contextual bandit problems.
\ourname contrasts with more classical approaches to exploration (like $\epsilon$-greedy or LinUCB; see \autoref{sec:related}), in which exploration strategies are constructed by expert algorithm designers.
These approaches often achieve provably good exploration strategies in the worst case, but are potentially overly pessimistic and are sometimes computationally intractable.

At training time (\autoref{sec:train}), \ourname simulates many contextual bandit problems from fully labeled synthetic data.
Using this data, in each round, \ourname is able to counterfactually simulate what would happen under all possible action choices.
We can then use this information to compute regret estimates for each action, which can be optimized using the AggreVaTe imitation learning algorithm~\citep{ross2014reinforcement}.
Our imitation learning strategy mirrors that of the meta-learning approach of \citet{bachman2017learning} in the active learning setting.
We present a simplified, stylized analysis of the behavior of \ourname to ensure that our cost function encourages good behavior (\autoref{sec:guarantees}).
Empirically, we use \ourname to train an exploration policy on only synthetic datasets and evaluate the resulting bandit performance across three hundred (simulated) contextual bandit tasks (\autoref{sec:tasks}), comparing to a number of alternative exploration algorithms, and showing the efficacy of our approach (\autoref{sec:comparison}).

%% file: approach.tex
%
\label{sec:cb}

Contextual bandits is a model of interaction in which an agent chooses actions (based on contexts) and receives immediate rewards for that action alone.
For example, in a simplified news personalization setting, at each time step $t$, a user arrives and the system must choose a news article to display to them.
Each possible news article corresponds to an action $a$, and the user corresponds to a context $x_t$.
After the system chooses an article $a_t$ to display, it can observe, for instance, the amount of time that the user spends reading that article, which it can use as a reward $r_t(a_t)$.
The goal of the system is to choose articles to display that maximize the cumulative sum of rewards, but it has to do this without ever being able to know what the reward would have been had it shown a different article $a'_t$.

Formally, we largely follow the setup and notation of \citet{agarwal14taming}.
Let $\cX$ be an input space of contexts (users) and $[K] = \{ 1, \dots, K \}$ be a finite action space (articles).
We consider the statistical setting in which there exists a fixed but unknown distribution $\cD$ over pairs $(x, \vec r) \in \cX \times [0,1]^{K}$, where $\vec r$ is a vector of rewards (for convenience, we assume all rewards are bounded in $[0,1]$).
In this setting, the world operates iteratively over rounds $t = 1, 2, \dots$.
Each round $t$:
\begin{enumerate}[nolistsep,noitemsep]
  \item The world draws $(x_t,\vec r_t) \sim \cD$ and reveals context $x_t$.
  \item The agent (randomly) chooses action $a_t \in [K]$ based on $x_t$, and observes reward $r_t(a_t)$.
\end{enumerate}
The goal of an algorithm is to maximize the cumulative sum of rewards over time.
Typically the primary quantity considered is the \emph{average regret} of a sequence of actions $a_1, \dots, a_T$ to the behavior of the best possible function in a prespecified class $\cF$:
\setlength{\belowdisplayskip}{0pt} \setlength{\belowdisplayshortskip}{0pt}
\setlength{\abovedisplayskip}{0pt} \setlength{\abovedisplayshortskip}{0pt}  
\begin{align} \label{eq:value}
  \reg(a_1, \dots, a_T) &= \max_{f \in \cF} \frac 1 T \sum_{t=1}^T \Big[ r_t(f(x_t)) - r_t(a_t)  \Big]
\end{align}
An agent is call \emph{no-regret} if its average regret is zero in the limit of large $T$.

\subsection{Policy Optimization over Fixed Histories} \label{sec:polopt}

To produce a good agent for interacting with the world, we assume access to a function class $\cF$ and to an \emph{oracle policy optimizer} for that function class.
For example, $\cF$ may be a set of single layer neural networks mapping user features (e.g., IP, browser, etc.) $x \in \cX$ 
to predicted rewards for actions (articles) $a \in [K]$, where $K$ is the total number of actions.
Formally, the observable record of interaction resulting from round $t$ is the tuple $(x_t, a_t, r_t(a_t), p_t(a_t)) \in \cX \times [K] \times [0,1] \times [0,1]$, where $p_t(a_t)$ is the probability that the agent chose action $a_t$, and the full history of interaction is $h_t = \langle (x_i, a_i, r_i(a_i), p_i(a_i)) \rangle_{i=1}^t$.
The oracle policy optimizer, $\polopt$, takes as input a \emph{history} of user interactions with the news recommendation system and outputs an $f \in \cF$ with low expected regret.

A standard example of a policy optimizer is to combine inverse propensity scaling (IPS) with a regression algorithm \citep{dudik2011efficient}.
Here, given a history $h$, each tuple $(x, a, r, p)$ in that history is mapped to a multiple-output regression example.
The input for this regression example is the same $x$; the output is a vector of $K$ costs, all of which are zero except the $a^{\text{th}}$ component, which takes value $r/p$.
For example, if the agent chose to show to user $x$ article $3$, made that decision with $80\%$ probability, and received a reward of $0.6$, then the corresponding output vector would be $\langle 0, 0, 0.75, 0, \dots, 0 \rangle$.
This mapping is done for all tuples in the history, and then a supervised learning algorithm on the function class $\cF$ is used to produce a low-regret regressor $f$. This is the function returned by the underlying policy optimizer.

IPS has this nice property that it is an unbiased estimator; unfortunately, it tends to have large variance especially when some probabilities $p$ are small.
In addition to IPS, there are several standard policy optimizers that mostly attempt to reduce variance while remaining unbiased: the direct method (which estimates the reward function from given data and
uses this estimate in place of actual reward), the double-robust estimator, and multitask regression.
In our experiments, we use the direct method because we found it best on average, but in principle any could be used.

\subsection{Test Time Behavior of \ourname} \label{sec:testtime}

In order to have an effective approach to the contextual bandit problem, one must be able to both optimize a policy based on historic data and make decisions about how to explore.
After all, in order for the example news recommendation system to learn whether a particular user is interested in news articles on some topic is to try showing such articles to see how the user responds (or to generalize from related articles or users).
The exploration/exploitation dilemma is fundamentally about long-term payoffs: is it worth trying something potentially suboptimal \emph{now} in order to learn how to behave better in the future?
A particularly simple and effective form of exploration is $\ep$-greedy: given a function $f$ output by \polopt, act according to $f(x)$ with probability $(1-\ep)$ and act uniformly at random with probability $\ep$.
Intuitively, one would hope to improve on such a strategy by taking more (any!) information into account; for instance, basing the probability of exploration on $f$'s uncertainty.

Our goal in this paper is to \emph{learn} how to explore from experience.
The training procedure for \ourname will use offline supervised learning problems to learn an \emph{exploration policy} $\pi$, which takes \emph{two inputs}: a function $f \in \cF$ and a context $x$, and outputs an action.
In our example, $f$ will be the output of the policy optimizer on all historic data, and $x$ will be the current user.
This is used to produce an agent which interacts with the world, maintaining an initially empty history buffer $h$, as:
\begin{enumerate}[nolistsep,noitemsep]
\item The world draws $(x_t,\vec r_t) \sim \cD$ and reveals context $x_t$.
\item The agent computes $f_t \gets \polopt(h)$ and a greedy action $\tilde a_t = \pi(f_t,x_t)$.
\item The agent plays $a_t = \tilde a_t$ with probability $(1-\mu)$, and $a_t$ uniformly at random otherwise.
\item The agent observes $r_t(a_t)$ and appends $(x_t, a_t, r_t(a_t), p_t)$ to the history $h$,
\item[] \quad where $p_t = \mu/K$ if $a_t \neq \tilde a_t$; and $p_t = 1-\mu+\mu/K$ if $a_t=\tilde a_t$.
\end{enumerate}
Here, $f_t$ is the function optimized on the historical data, and $\pi$ uses it and $x_t$ to choose an action.
Intuitively, $\pi$ might choose to use the prediction $f_t(x_t)$ most of the time, unless $f_t$ is quite uncertain on this example, in which case $\pi$ might choose to return the second (or third) most likely action according to $f_t$.
The agent then performs a small amount of additional $\mu$-greedy-style exploration: most of the time it acts according to $\pi$ but occasionally it explores some more.
In practice (\autoref{sec:experiments}), we find that setting $\mu=0$ is optimal in aggregate, but non-zero $\mu$ is necessary for our theory (\autoref{sec:guarantees}).

\subsection{Training \ourname by Imitation Learning} \label{sec:train}

The meta-learning challenge is: how do we learn a good exploration policy $\pi$?
We assume we have access to \emph{fully labeled} data on which we can train $\pi$; this data must include context/reward pairs, but where the reward for \emph{all} actions is known.
This is a weak assumption: in practice, we use purely synthetic data as this training data; one could alternatively use any fully labeled classification dataset (this is inspired by \citet{beygelzimer2009offset}).
Under this assumption about the data, it is natural to think of $\g$'s behavior as a sequential decision making problem in a simulated setting, for which a natural class of learning algorithms to consider are imitation learning algorithms \citep{daume2009search,ross2010reduction,ross2014reinforcement,chang2015learning}.\footnote{In other work on meta-learning, such problems are often cast as full \emph{reinforcement-learning} problems. We opt for imitation learning instead because it is computationally attractive and effective when a simulator exists.}
Informally, at training time, \ourname will treat one of these synthetic datasets as if it were a contextual bandit dataset.
At each time step $t$, it will compute $f_t$ by running $\polopt$ on the historical data, and then ask: for \emph{each} action, what would the long time reward look like if I were to take this action.
Because the training data for \ourname is fully labeled, this can be evaluated for each possible action, and a policy $\pi$ can be learned to maximize these rewards.

Importantly, we wish to train $\g$ using one set of tasks (for which we have fully supervised data on which to run simulations) and apply it to wholly different tasks (for which we only have bandit feedback).
To achieve this, we allow $\g$ to depend representationally on $f_t$ in arbitrary ways: for instance, it might use features that capture $f_t$'s uncertainty on the current example (see \autoref{sec:features} for details).
We additionally allow $\g$ to depend in a \emph{task-independent} manner on the history (for instance, which actions have not yet been tried): it can use features of the actions, rewards and probabilities in the history but \emph{not} depend directly on the contexts $x$.
This is to ensure that $\g$ only learns to explore and not also to solve the underlying task-dependent classification problem.

More formally, in imitation learning, we assume training-time access to an \emph{expert}, $\gstar$, whose behavior we wish to learn to imitate at test-time.
From this, we can define an optimal reference policy $\gstar$, which effectively ``cheats'' at training time by looking at the true labels.
The learning problem is then to estimate $\g$ to have as similar behavior to $\gstar$ as possible, but without access to those labels.
Suppose we wish to learn an exploration policy $\g$ for a contextual bandit problem with $K$ actions.
We assume access to $M$ supervised learning datasets $S_1, \dots, S_M$, where each $S_m = \{ (x_1, \vec r_1), \dots, (x_{N_m}, \vec r_{N_m}) \}$ of size $N_m$, where each $x_n$ is from a (possibly different) input space $\cX_m$ and the reward vectors are all in $[0,1]^K$.%
We wish to learn an exploration policy $\g$ with maximal reward: \emph{therefore}, $\g$ should imitate a $\gstar$ that always chooses its action optimally.

We additionally allow $\g$ to depend on a \emph{very small} amount of fully labeled data from the task at hand, which we use to allow $\g$ to calibrate $f_t$'s predictions.Because $\g$ needs to learn to be task independent, we found that if $f_t$s were uncalibrated, it was very difficult for $\g$ to generalize well to unseen tasks. In our experiments we use only $30$ fully labeled examples, but alternative approaches to calibrating $f_t$ that do not require this data would be ideal.

\begin{algorithm}[t]
  \caption{\algorithmname{\ourname}$($supervised training sets $\{S_m\}$, hypothesis class $\cF$, exploration rate $\mu=0.1$, number of validation examples $N_{\textit{Val}}=30)$, feature extractor $\Phi$}
  \label{alg:train}
  \begin{algorithmic}[1]
    \FOR{round $n = 1, 2, \dots, N$}
      \STATE initialize meta-dataset $D = \{\}$ and choose dataset $S$ at random from $\{ S_m \}$
      \STATE partition and permute $S$ randomly into train $\textit{Tr}$ and validation $\textit{Val}$ where $\card{\textit{Val}} = N_\textit{Val}$ \label{train:partition}
      \STATE set history $h_0 = \{\}$
      \FOR{round $t = 1, 2, \dots, \card{\textit{Tr}}$}
        \STATE let $(x_t, \vec r_t) = \textit{Tr}_t$
        \FOR{each action $a = 1, \dots, K$}
          \STATE optimize $f_{t,a} = \polopt(\cF, h_{t-1} \oplus (x_t, a, $ \\ $r_t(a), 1$-$(K$-$1)\mu))$ on augmented history\label{train:newhistory}
          \STATE roll-out: estimate $\hat\rho_a$, the value of $a$, using $r_t(a)$ and a roll-out policy $\pi^\textrm{out}$ \label{train:estimate} %
        \ENDFOR
        \STATE compute $f_t = \polopt(\cF, h_{t-1})$ \label{train:ft}
        \STATE aggregate $D \gets D \oplus (\Phi(f_t, x_t, h_{t-1}, \textit{Val}), \langle \hat\rho_1, $\\$\dots, \hat\rho_K\rangle)$
        \label{train:aggregate}
        \STATE roll-in: $a_t \sim \frac \mu K \1_K + (1-\mu) \g_{n-1}(f_t, x_t)$ with probability $p_t$, $\1$ is an indicator function \label{train:rollin}
        \STATE append history $h_t \gets h_{t-1} \oplus (x_t, a_t, r_t(a_t), p_t)$
      \ENDFOR
      \STATE update $\g_n = \textsc{Learn}(D)$ \label{train:update}
    \ENDFOR
    \RETURN $\{ \g_n \}_{n=1}^N$%
  \end{algorithmic}
\end{algorithm}

The imitation learning algorithm we use is AggreVaTe \citep{ross2014reinforcement} (closely related to DAgger \citep{ross2010reduction}), and is instantiated for the contextual bandits meta-learning problem in \autoref{alg:train}. AggreVaTe learns to choose actions to minimize the cost-to-go of the expert 
rather than the zero-one classification loss of mimicking its actions. On the first iteration AggreVaTe collects data 
by observing the expert perform the task, and in each trajectory, at time $t$, explores an action $a$ in state $s$, and 
observes the cost-to-go $Q$ of the expert after performing this action. 

Each of these steps generates a cost-weighted training example $(s, t, a, Q)$ and AggreVaTe trains a policy $\pi_1$ to 
minimize the expected cost-to-go on this dataset.  At each following iteration $n$, AggreVaTe collects data through 
interaction with the learner as follows: for each trajectory, begin by using the current learner's policy $\pi_n$ to 
perform the task, interrupt at time $t$, explore a roll-in action $a$ in the current state $s$, after which control is 
provided back to the expert to continue up to time-horizon $T$. This results in new examples of the cost-to-go 
(roll-out value) of the expert $(s, t, a, Q)$, under the distribution of states visited 
by the current policy $\pi_n$.  This new data is aggregated with all previous data to train the next policy $\pi_{n+1}$; 
more generally, this data can be used by a no-regret online learner to update the policy and obtain $\pi_{n+1}$. This is 
iterated for some number of iterations $N$ and the best policy found is returned. AggreVaTe optionally allow the algorithm 
to continue executing the expert’s actions with small probability $\beta$, instead of always executing $\pi_n$, up to the 
time step $t$ where an action is explored and control is shifted to the expert.

\ourname operates in an iterative fashion, starting with an arbitrary $\g$ and improving it through interaction with an expert.
Over $N$ rounds, \ourname selects random training sets and simulates the test-time behavior on that training set.
The core functionality is to generate a number of states $(f_t, x_t)$ on which to train $\g$, and to use the supervised data to estimate the value of every action from those states.
\ourname achieves this by sampling a random supervised training set and setting aside some validation data from it (\autoref{train:partition}).
It then simulates a contextual bandit problem on this training data; at each time step $t$, it tries \emph{all} actions and ``pretends'' like they were appended to the current history (\autoref{train:newhistory}) on which it trains a new policy and evaluates it's \textbf{roll-out value} (\autoref{train:estimate}, described below).
This yields, for each $t$, a new training example for $\g$, which is added to $\g$'s training set (\autoref{train:aggregate}); the features for this example are features of the classifier based on true history (\autoref{train:ft}) (and possibly statistics of the history itself), with a label that gives, for each action, the corresponding value of that action (the $\rho_a$s computed in \autoref{train:estimate}).
\ourname then must commit to a \textbf{roll-in action} to \emph{actually} take; it chooses this according to a roll-in policy (\autoref{train:rollin}), described below.

The two key questions are: how to choose roll-in actions and how to evaluate roll-out values.

\textbf{Roll-in actions.}\quad
The distribution over states visited by \ourname depends on the actions taken, and in general it is good to have that distribution match what is seen at test time as closely as possible.
This distribution is determined by a \emph{roll-in} policy (\autoref{train:rollin}), controlled in \ourname by exploration parameter $\mu \in [0,1/K]$.
As $\mu \rightarrow 1/K$, the roll-in policy approaches a uniform random policy; as $\mu \rightarrow 0$, the roll-in policy becomes deterministic.
When the roll-in policy does not explore, it acts according to $\g(f_t, .)$.

\textbf{Roll-out values.}\quad
The ideal value to assign to an action (from the perspective of the imitation learning procedure) is that total reward (or advantage) that would be achieved in the long run if we took this action and then behaved according to our final learned policy.
Unfortunately, during training, we do not yet know the final learned policy. Thus, a surrogate roll-out policy $\pi^\textrm{out}$ 
is used instead. A convenient, and often computationally efficient alternative, is to evaluate the value assuming all future actions were taken by the expert \citep{langford2005relating,daume2009search,ross2014reinforcement}. In our setting, at any time step $t$, the expert has access to the fully supervised reward vector $\vec r_t$ for the context
$\vec x_t$. When estimating the roll-out value for an action $a$, the expert will return the true reward value for this action $r_t(a)$ and  we use this as our estimate for the roll-out value.

%% file: guarantees.tex
%

We analyze \ourname, showing that the no-regret property of \algorithmname{AggreVaTe} can be leveraged in our meta-learning setting for learning contextual bandit exploration. In particular, we first relate the regret of the learner in line 16 to the overall regret of $\pi$. This will show that, \emph{if} the underlying classifier improves sufficiently quickly, \ourname will achieve sublinear regret. We then show that for a specific choice of underlying classifier (\algorithmname{Banditron}), this is achieved.

\ourname is an instantiation of \algorithmname{AggreVaTe} \citep{ross2014reinforcement}; as such, it inherits \algorithmname{AggreVaTe}'s regret guarantees.
Let $\hat\ep_{\textit{class}}$ denote the empirical minimum expected cost-sensitive classification regret achieved by policies in the class $\Pi$ on all the data over the $N$ iterations of training when compared to the Bayes optimal regressor, for $U(T)$ the uniform distribution over $\{1, \dots, T\}$, $d^t_\pi$ the distribution of states at time $t$ induced by executing policy $\pi$, and $Q^\star$ the cost-to-go of the expert: 

\begin{equation*}
\begin{aligned}
    \hat\ep_{\textit{class}}(T) = \min_{\pi \in \Pi}\frac{1}{N} \hat\Ep_{t\sim U(T),s\sim d^t_{\pi_i}} \sum_{i=1}^{N} \Big[Q^\star_{T-t+1}(s, \pi) \\
    - \min_a Q ^\star_{T-t+1}(s, a)\Big]
\end{aligned}
\end{equation*}

\begin{theorem}[Thm 2.2 of \citet{ross2014reinforcement}, adapted] \label{thm:regret}
  After $N$ rounds in the parameter-free setting, if a \textsc{Learn} (\autoref{train:update}) is no-regret algorithm, then as $N \rightarrow \infty$, with probability $1$, it holds that
  $J(\bar \g) \leq J(\gstar) + 2 T \sqrt{K \hat\ep_{\textit{class}}(T)}$,
  where $J(\cdot)$ is the reward of the exploration policy, $\bar \g$ is the average policy returned, and $\hat\ep_{\textit{class}}(T)$ is the average regression regret for each $\g_n$ accurately predicting $\hat\rho$.
\end{theorem}

This says that if we can achieve low regret at the problem of learning $\g$ on the training data it observes (``$D$'' in \ourname), i.e. $\hat\ep_{\textit{class}}(T)$ is small, then this translates into low regret in the contextual-bandit setting.

At first glance this bound looks like it may scale linearly with $T$. However, the bound in~\autoref{thm:regret} is dependent on $\hat\ep_{\textit{class}}(T)$.
Note however, that $s$ is a combination of the context vector $x_t$ and the classification function $f_t$.
As $T\rightarrow \infty$, one would hope that $f_t$ improves significantly and $\hat\ep_{\textit{class}}(T)$ decays quickly. 
Thus, sublinear regret may still be achievable when $f$ learns sufficiently quickly as a function of $T$.
For instance, if $f$ is optimizing a strongly convex loss function, online gradient descent achieves a regret guarantee of $O(\frac{\log T}{T})$ (e.g., Theorem 3.3 of~\citet{hazan2016introduction}), potentially leading to a regret for \ourname of $O(\sqrt{(\log T)/{T}})$.

The above statement is informal (it does not take into account the interaction between learning $f$ and $\pi$).
However, we can show a specific concrete example: we analyze \ourname's test-time behavior when the underlying learning algorithm is \algorithmname{Banditron}.
\algorithmname{Banditron} is a variant of the multiclass Perceptron that operates under bandit feedback.
Details of this analysis (and proofs, which directly follow the original \algorithmname{Banditron} analysis) are given in \autoref{apx:banditron}; here we state the main result.
Let $\ga_t = \Pr[r_t(\g(f_t, x_t) = 1) | x_t] - \Pr[r_t(f_t(x_t)) = 1 | x_t]$ be the edge of $\g(f_t, .)$ over $f$, and let $\Gamma = \frac 1 T \sum_{t=1}^T \Ep \frac 1 {1 + K\ga_t}$ be an overall measure of the edge. For instance if $\g$ simply returns $f$'s prediction, then all $\ga_t=0$ and $\Ga=1$.
We can then show the following:

\begin{toappendix}
\section{Stylized test-time analysis for Banditron: Details} \label{apx:banditron}
  
The \algorithmname{Banditron\ourname} algorithm is specified in \autoref{alg:polelim}. The is exactly the same as the typical test time behavior, except it uses a \algorithmname{Banditron}-type strategy for learning the underlying classifier $f$ in the place of \polopt. \algorithmname{PolicyEliminationMeta} takes as arguments: $\g$ (the learned exploration policy) and $\mu \in (0, 1/(2K))$ an added uniform exploration parameter.
The \algorithmname{Banditron} learns a linear multi-class classifier parameterized by a weight matrix of size $K \times D$, where $D$ is the input dimensionality.
The \algorithmname{Banditron} assumes a pure multi-class setting in which the reward for one (``correct'') action is 1 and the reward for all other actions is zero.

At each round $t$, a prediction $\hat a_t$ is made according to $f_t$ (summarized by $W^t$).
We then define an exploration distribution that ``most of the time'' acts according to $\g(f_t, .)$, but smooths each action with $\mu$ probability.
The chosen action $a_t$ is sampled from this distribution and a binary reward is observed.
The weights of the \algorithmname{Banditron} are updated according to the \algorithmname{Banditron} update rule using $\tilde U^t$.

\begin{algorithm}[h]
  \caption{\algorithmname{Banditron\ourname}$(g, \mu)$}
  \label{alg:polelim}
  \begin{algorithmic}[1]
    \STATE initialize $W^1 = \vec 0 \in \R^{K \times D}$
    \FOR{rounds $t = 1 \dots T$:}
    \STATE observe $x_t \in \R^D$
    \STATE compute $\hat a_t = f_t(x_t) = \argmax_{k \in K} \big( W^t x_t \big)_k$
    \STATE define $Q^\mu(a) = \mu + (1-K\mu) \Ind[a = \g(W^t, x_t)]$
    \STATE sample $a_t \sim Q^\mu$
    \STATE observe reward $r_t(a_t) \in \{0,1\}$
    \STATE define $\tilde U^t \in \R^{K \times D}$ as: \\
      \quad $\tilde U^t_{a,\cdot} = x_t \left( \frac {\Ind[r_t(a_t)=1] \Ind[a_t=a]} {Q^\mu(a)} - \Ind[\hat a_t=a] \right)$
    \STATE update $W^{t+1} = W^t + \tilde U^t$
    \ENDFOR
  \end{algorithmic}
\end{algorithm}

The \emph{only} difference between \algorithmname{Banditron\ourname} and the original \algorithmname{Banditron} is the introduction of $\g$ in the sampling distribution. The original algorithm achieves the following mistake bound shown below, which depends on the notion of multi-class hinge-loss.
In particular, the hinge-loss of $W$ on $(x,\vec r)$ is $\ell(W,(x,\vec r)) = \max_{a \neq a^\star} \max \big\{ 0, 1 - (Wx)_{a^\star} + (Wx)_a \big\}$, where $a^\star$ is the $a$ for which $r(a)=1$. The overall hinge-loss $L$ is the sum of $\ell$ over the sequence of examples.

\begin{theorem}[Thm.~1 and Corr. 2 of \citet{kakade2008banditron}] \label{thm:banditron}
  Assume that for the sequence of examples, $(x_1, \vec r_1), (x_2, \vec r_2), \dots, (x_T, \vec r_T)$, we have, for all $t$, $\norm{x_t} \leq 1$.
  Let $W^\star$ be any matrix, let $L$ be the cumulative hinge-loss of $W^\star$, and let $D = 2 \norm{W^\star}_F^2$ be the complexity of $W^\star$.
  The number of mistakes $M$ made by the \algorithmname{Banditron} satisfies
  \begin{equation}
    \Ep M \leq L + K\mu T + 3 \max \left\{ \frac D \mu, \sqrt{DTK\mu} \right\} + \sqrt{DL/\mu}
  \end{equation}
  where the expectation is taken with respect to the randomness of the algorithm.
  Furthermore, in a low noise setting (there exists $W^\star$ with fixed complexity $d$ and loss $L \leq O(\sqrt{DKT})$), then by setting $\mu = \sqrt{D/(TK)}$, we obtain $\Ep M \leq O(\sqrt{KDT})$.
\end{theorem}

We can prove an analogous result for \algorithmname{Banditron\ourname}.
The key quantity that will control how much $\g$ improves the execution of \algorithmname{Banditron\ourname} is how much $\g$ improves on $f_t$ when $f_t$ is wrong.
In particular, let $\ga_t = \Pr[r_t(\g(f_t, x_t) = 1) | x_t] - \Pr[r_t(f_t(x_t)) = 1 | x_t]$ be the edge of $\g(f_t, .)$ over $f$, and let $\Ga = \frac 1 T \sum_{t=1}^T \Ep \frac 1 {1 + K\ga_t}$ be an overall measure of the edge. (If $\g$ does nothing, then all $\ga_t=0$ and $\Ga=1$.)
Given this quantity, we can prove the following \autoref{thm:us}.

\begin{proof}[sketch]
  The proof is a small modification of the original proof of \autoref{thm:banditron}.
  The only change is that  in the original proof, the following bound is used:
  $\Ep_t \normsmall{\tilde U^t}^2 / \normsmall{x_t}^2 = 1 + 1/\mu \leq 2/\mu$.
  We use, instead:
  $\Ep_t \normsmall{\tilde U^t}^2 / \normsmall{x_t}^2
  \leq 1 + \Ep_t \frac 1 {\mu + \ga_t} \leq \frac {2 \Ep_t \frac 1 {1 + \ga_t}} \mu$.
  The rest of the proof goes through identically.
\end{proof}
\end{toappendix}

\begin{theorem} \label{thm:us}
  Assume that for the sequence of examples, $(x_1, \vec r_1), (x_2, \vec r_2), \dots, (x_T, \vec r_T)$, we have, for all $t$, $\norm{x_t} \leq 1$.
  Let $W^\star$ be any matrix, let $L$ be the cumulative hinge-loss of $W^\star$, let $\mu$ be a uniform exploration probability, and let $D = 2 \norm{W^\star}_F^2$ be the complexity of $W^\star$.
  Assume that $\Ep \ga_t \geq 0$ for all $t$.
  Then the number of mistakes $M$ made by \ourname with \algorithmname{Banditron} as \textsc{PolOpt} satisfies:
  \begin{equation}
    \Ep M \leq L + K\mu T + 3 \max \left\{ {D\Ga} / \mu, \sqrt{DTK\Ga\mu} \right\} + \sqrt{DL\Ga/\mu}
  \end{equation}
  where the expectation is taken with respect to the randomness of the algorithm.
\end{theorem}

Note that under the assumption $\Ep \gamma_t \geq 0$ for all $t$, we have $\Gamma \leq 1$.
The analysis gives the same mistake bound for \algorithmname{Banditron} but with the additional factor of $\Gamma$, hence this result improves upon the standard \algorithmname{Banditron} analysis only when $\Gamma < 1$.

This result is highly stylized and the assumption that $\Ep \ga_t \geq 0$ is overly strong. This assumption ensures that $\g$ never decreases the probability of a ``correct'' action.
It does, however, help us understand the behavior of \ourname, qualitatively:
First, the quantity that matters in \autoref{thm:us}, $\Ep_t \ga_t$ is (in the 0/1 loss case) exactly what \ourname is optimizing: the expected improvement for choosing an action against $f_t$'s recommendation.
Second, the benefit of using $\g$ within \algorithmname{Banditron} is a \emph{local} benefit: because $\g$ is trained with expert rollouts, as discussed in \autoref{sec:guarantees}, the primary improvement in the analysis is to ensure that $\g$ does a better job predicting (in a single step) than $f_t$ does.
An obvious open question is whether it is possible to base the analysis on the \emph{regret} of $\g$ (rather than its error) and whether it is possible to extend beyond the simple \algorithmname{Banditron} setting.

%% file: experiments.tex
%

\label{sec:experiments}

\begin{figure*}[t]
  \centering
  \includegraphics[height=4.5cm,trim={10 10 0 5},clip]{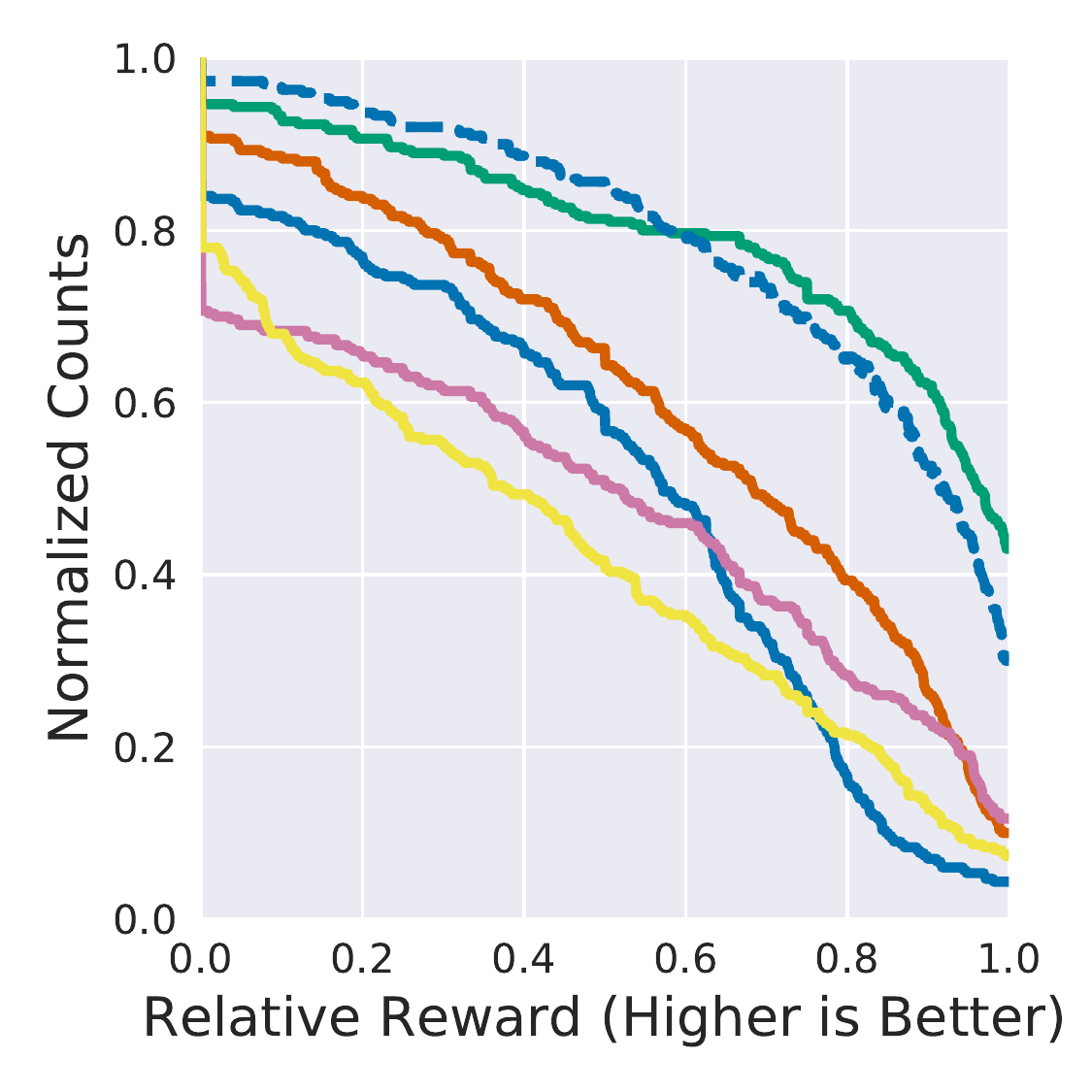}\hfill
  \includegraphics[height=4.5cm,trim={0 0 35 30},clip]{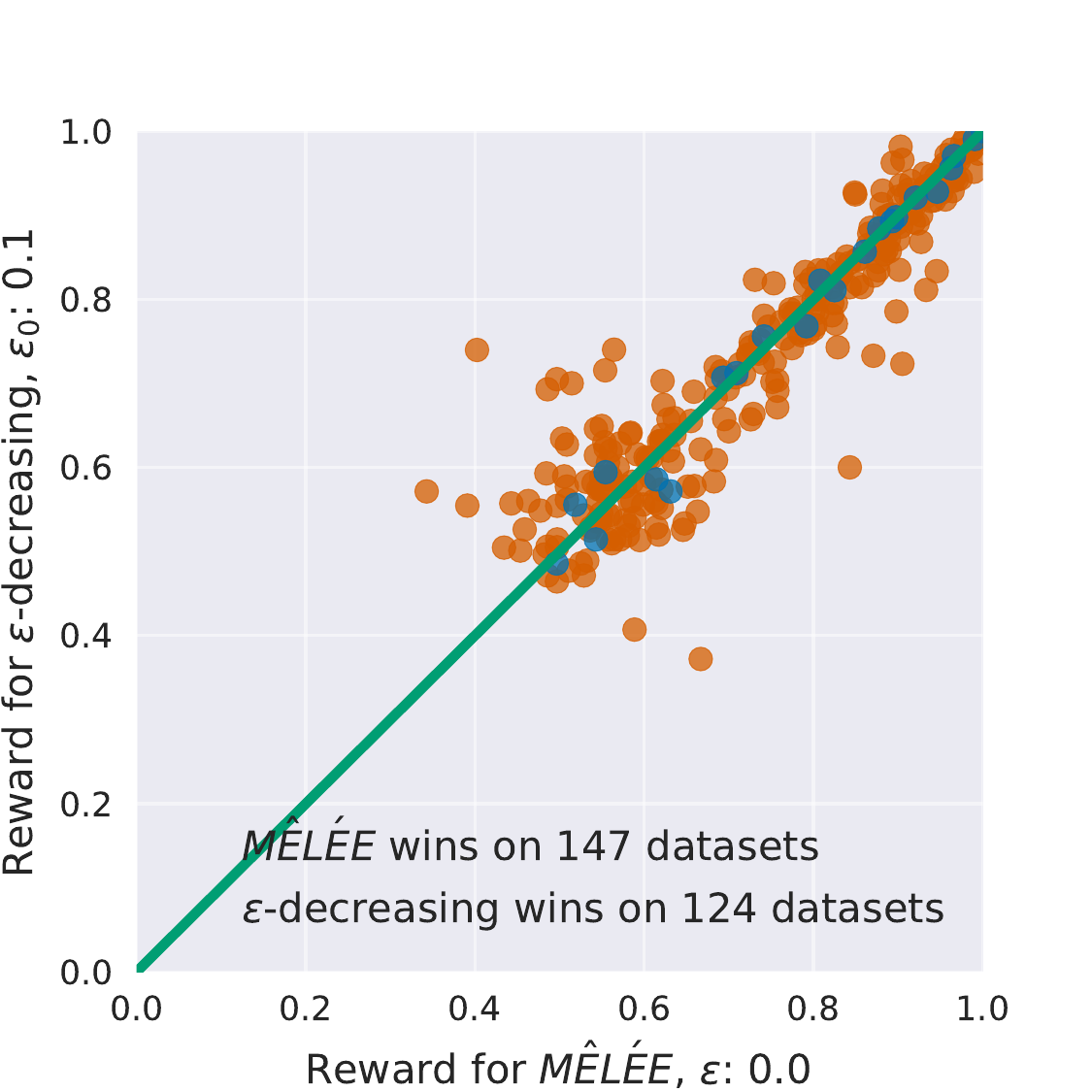}\hfill
  \includegraphics[height=4.5cm,trim={10 15 10 10},clip]{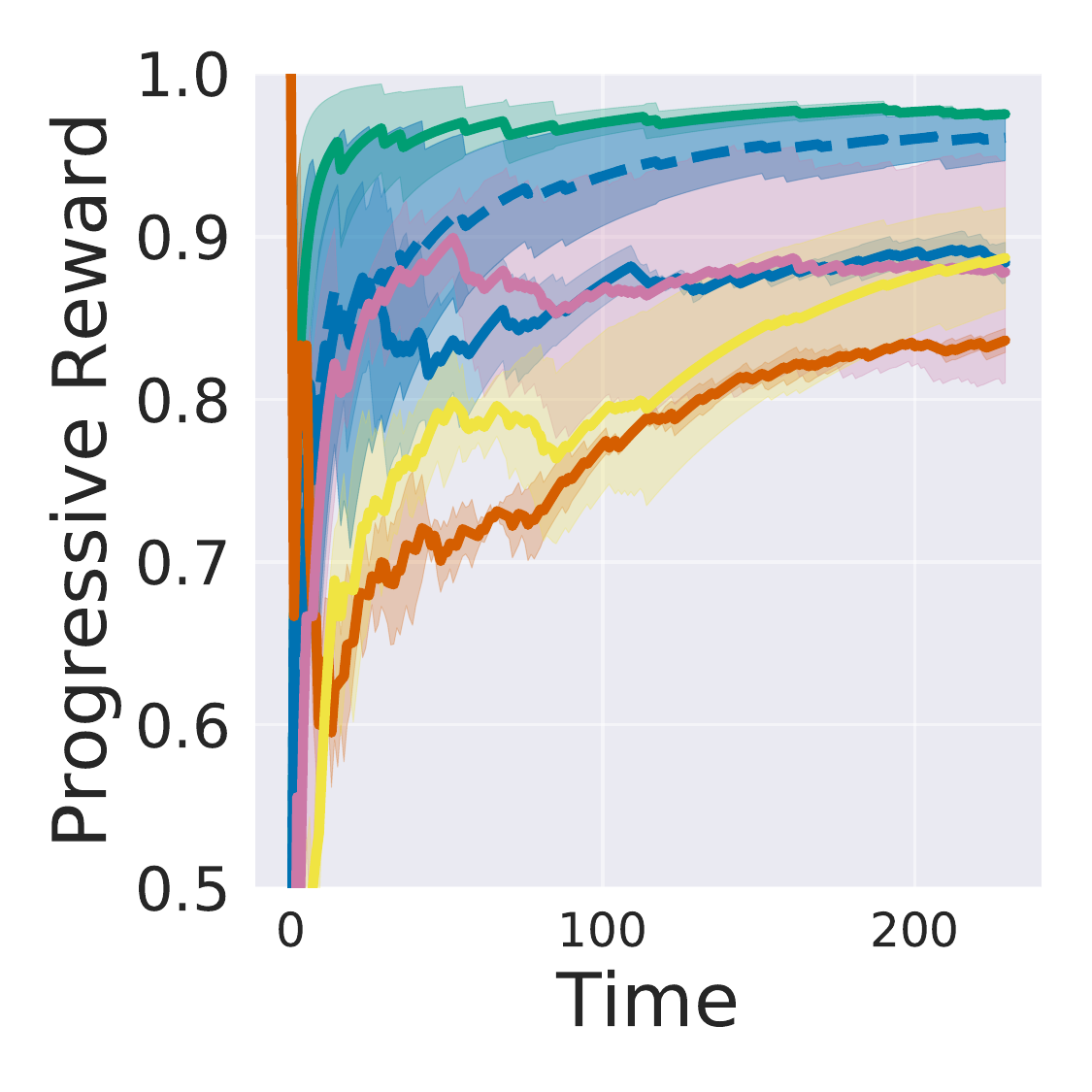}
  \includegraphics[height=4.5mm,trim={590 10 1150 10},clip]{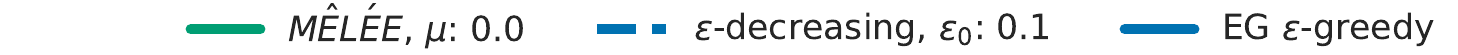}
  \includegraphics[height=4.5mm,trim={1010 10 50 10},clip]{fig/legend.pdf}
  \caption{Comparison of algorithms on $300$ 
      classification problems. \textbf{(Left)}
      Comparison  of  all  exploration algorithms  using  the  empirical 
      cumulative distribution function of the relative progressive validation 
      return $G$ (upper-right is optimal). The curves for $\ep$-decreasing \& $\ep$-greedy coincide.
      \textbf{(Middle)} Comparison of \ourname to the second best 
      performing exploration algorithm ($\ep$-decreasing), every data point represents 
      one of the $300$ datasets, x-axis shows the reward of $G(\textrm{\ourname})$, y-axis show 
      the reward of $G(\ep$-decreasing$)$, and red dots represent 
      statistically significant runs. \textbf{(Right)} A representative learning curve on dataset \#1144.}
  \label{fig:overall}
\end{figure*}

\begin{figure}
\footnotesize\vspace{-1.5em}
\begin{center}
    \includegraphics[width=.35\textwidth]{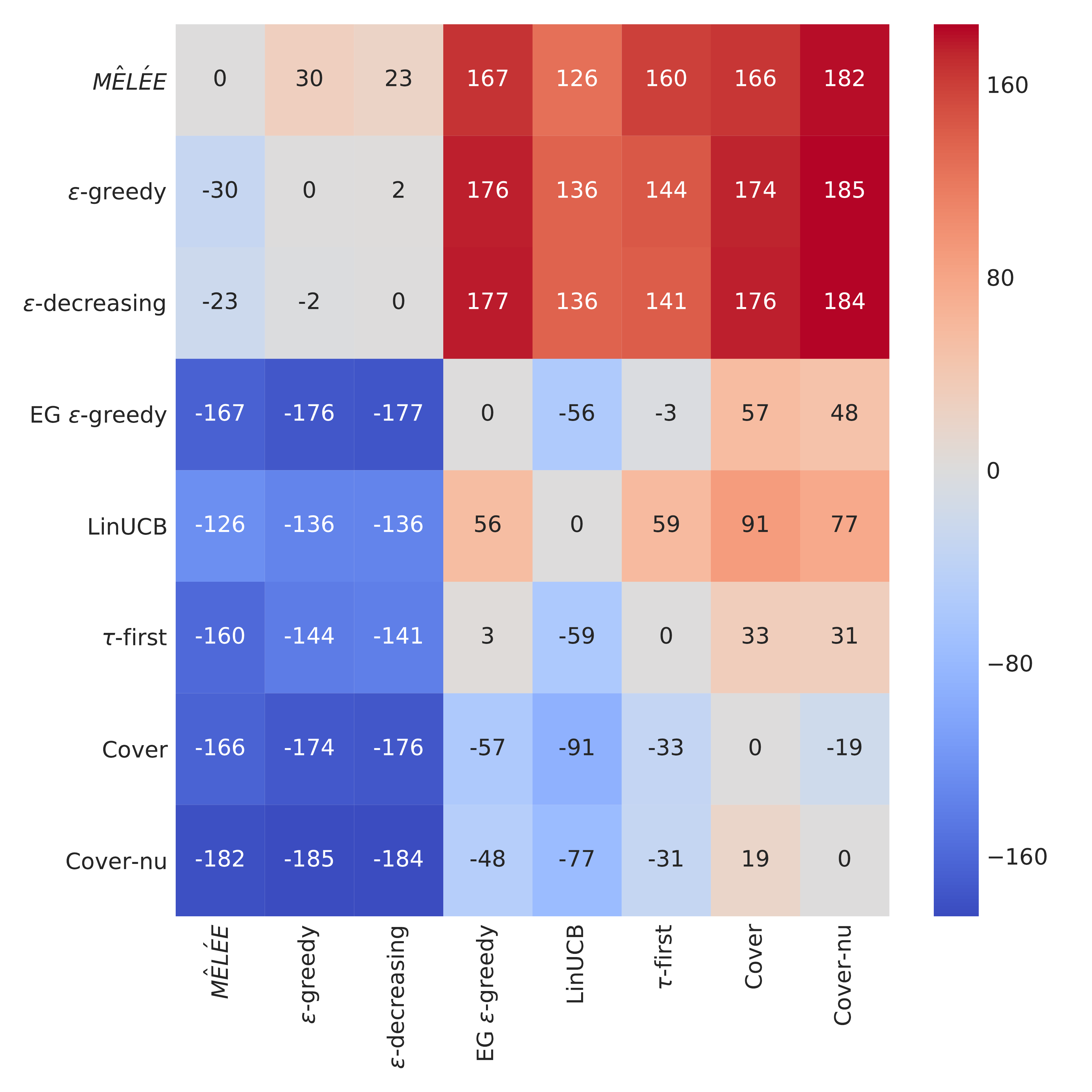}
\end{center}
\caption{\footnotesize Win statistics: each (row, column) entry shows the number of times the row algorithm won 
    against the column, minus the number of losses.
\label{tab:winlose}}\vspace{-2em}
\end{figure}

\begin{figure}
  \centering
  \includegraphics[width=0.6\linewidth,trim={0 0 35 35},clip]{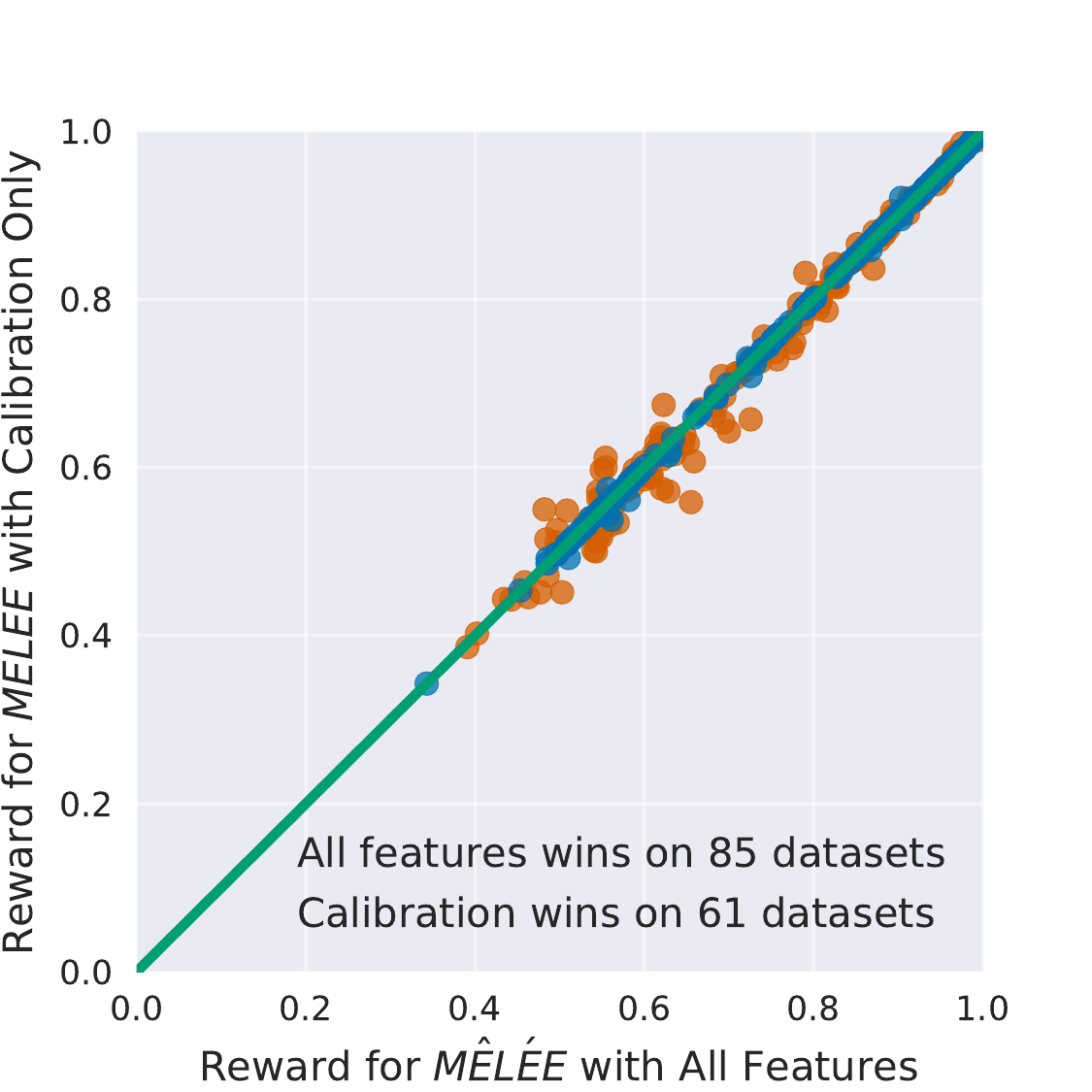} \vspace{-1em}
  \caption{\footnotesize Comparison of training \ourname with all the features (\autoref{sec:features}, y-axis) vs 
  training using only the calibrated prediction probabilities (x-axis). \ourname gets 
  an additional leverage when using all the features.}\vspace{-4em}
  \label{fig:ablation}
  \vspace{3em}
\end{figure}

Our experimental setup operates as follows:
  Using a collection of synthetically generated classification problems, we train an exploration policy $\pi$ using \ourname (\autoref{alg:train}).
  This exploration policy learns to explore on the basis of calibrated probabilistic predictions from $f$ together with a predefined set of exploration features (\autoref{sec:features}).
  Once $\pi$ is learned and fixed, we follow the test-time behavior described in \autoref{sec:testtime} on a set of $300$ ``simulated'' contextual bandit problems, derived from standard classification tasks.
In all cases, the underlying classifier $f$ is a linear model trained with a policy optimizer that runs stochastic gradient descent.

We seek to answer two questions experimentally: (1) How 
does \ourname compare empirically to alternative (expert designed) exploration strategies?
(2) How important are the additional features used by \ourname in comparison to using calibrated probability 
predictions from $f$ as features?

\subsection{Training Details for the Exploration Policy} \label{sec:synthetic}

\textbf{Exploration Features.} \label{sec:features}
In our experiments, the exploration policy is trained based on features $\Phi$ (\autoref{alg:train}, \autoref{train:aggregate}).
These features are allowed to depend on the current classifier $f_t$, and on any part of the history \emph{except} the inputs $x_t$ in order to maintain task independence.
We additionally ensure that its features are independent of the \emph{dimensionality} of the inputs, so that $\g$ can generalize to datasets of arbitrary dimensions.
The specific features we use are listed below; these are largely inspired by \citet{konyushkova2017metaactive} but adapted and augmented to our setting.
The \textbf{features of $f_t$} that we use are:
\begin{enumerate*}[label={\alph*)},font={\bfseries}]
\item predicted probability $p(a_t | f_t, \vec x_t)$;
\item entropy of the predicted probability distribution;
\item a one-hot encoding for the predicted action $f_t(\vec x_t)$.
\end{enumerate*}
The \textbf{features of $h_{t-1}$} that we use are:
\begin{enumerate*}[label={\alph*)},font={\bfseries}]
\item current time step $t$;
\item normalized counts for all previous actions predicted so far;
\item average observed rewards for each action;
\item empirical variance of the observed rewards for each action in the history.
\end{enumerate*}
In our experiments, we found that it is essential to calibrate the predicted probabilities of the classifier $f_t$.
We use a very small held-out dataset, of size $30$, to achieve this.
We use Platt's scaling~\citep{Platt99probabilisticoutputs,Lin2007} method to calibrate 
the predicted probabilities. Platt's scaling works by fitting a logistic regression 
model to the classifier's predicted scores.

\textbf{Training Datasets.} In our experiments, we follow \citet{konyushkova2017metaactive} (and also~\citet{peters2014causal}, in a different setting) and train the exploration policy $\g$ only on \emph{synthetic data}.
This is possible because the exploration policy $\pi$ never makes use of $x$ explicitly 
and instead only accesses it via $f_t$'s behavior on it. We generate datasets
with uniformly distributed class conditional distributions. The datasets are always 
two-dimensional. Details are in \autoref{apx:synthetic}.

\subsection{Details of Synthetic Datasets} \label{apx:synthetic}
We generate datasets with uniformly distributed class conditional distributions.
We generate 2D datasets by first sampling a random variable representing the 
Bayes classification error. The Bayes error is sampled uniformly from the interval $0.0$ to $0.5$. 
Next, we generate a balanced dataset where the data for each class lies within a
unit rectangle and sampled uniformly. We overlap the sampling rectangular 
regions to generate a dataset with the desired Bayes error selected in the first
step.

\subsection{Implementation Details.}
Our implementation is based on scikit-learn~\citep{scikit-learn}. We fix the training time exploration parameter $\mu$ 
to $0.1$. %
We train the exploration policy $\g$ on $82$ synthetic datasets each of size $3000$ with uniform class conditional distributions, a total of $246k$ samples (\autoref{apx:synthetic}).
We train $\g$ using a linear classifier~\cite{Breiman:2001:RF:570181.570182} and set the hyper-parameters for the 
learning rate, and data scaling methods using three-fold cross-validation on the whole meta-training dataset.
For the classifier class $\cF$, we use a linear model trained with stochastic gradient descent. We standardize all features 
to zero mean and unit variance, or scale the features to lie between zero and one. To select between the two scaling 
methods, and tune the classifier's learning rate, we use three-fold cross-validation on a small fully supervised 
training set of size $30$ samples. The same set is used to calibrate the predicted probabilities of $f_t$.

\subsection{Evaluation Tasks and Metrics}  \label{sec:tasks} \label{sec:metric}

Following~\citet{bietti:hal-01708310}, we use a collection of $300$ binary 
classification datasets from
\url{openml.org} for evaluation; the precise list and download instructions is in \autoref{apx:listofdatasets}. These datasets cover a variety of different domains 
including text \& image processing, medical diagnosis, and sensory data.
We convert multi-class classification datasets into cost-sensitive 
classification problems by using a $0/1$ encoding.%
Given these fully supervised cost-sensitive multi-class datasets, we simulate 
the contextual bandit setting by only revealing the reward for 
the selected actions. 
For evaluation, we use progressive validation \citep{blum1999beating}, which is exactly computing the reward of the algorithm.
Specifically, to evaluate the performance of an 
exploration algorithm $\cal A$ on a dataset $S$ of size $n$, we compute the 
progressive validation return $G({\cal A}) = \frac{1}{n}\sum_{t=1}^{n} r_t(a_t)$ as the average reward up to $n$, where $a_t$ is the action chosen by the algorithm $\cal A$ and $r_t$ is 
the true reward vector. %

\begin{toappendix}
  \input{datasets}
\end{toappendix}

Because our evaluation is over $300$ datasets, we report aggregate results in two forms.
The simpler one is \textbf{Win/Loss Statistics:}
We compare two exploration methods on a given 
dataset by counting the number of statistically significant wins and losses.
An exploration algorithm $\cal A$ wins over another algorithm $\cal B$ if the 
progressive validation return $G({\cal A})$ is statistically significantly larger than $B$'s return  
$G({\cal B})$ at the $0.01$ level using a paired sample t-test.
We also report \textbf{cumulative distributions} of rewards for each algorithm. In particular, for a given 
relative reward value ($x \in [0,1]$), the corresponding CDF value for a given algorithm is the fraction of datasets on 
which this algorithm achieved reward at least $x$. We compute relative reward by Min-Max normalization. 
Min-Max normalization linearly transforms reward $y$ to $x = \frac{y-\textrm{min}}{\textrm{max}-\textrm{min}}$, where $\textrm{min}$ \& $\textrm{max}$ are the minimum \& maximum rewards among all exploration algorithms. 

\subsection{Baseline Exploration Algorithms} \label{sec:baselines}

Our experiments aim to determine how \ourname compares to other standard exploration strategies.
In particular, we compare to:
\begin{description}[nolistsep,noitemsep,leftmargin=1em]
\item[$\ep$-greedy:] With probability $\ep$, explore uniformly at random; with probability $1-\ep$ act greedily according to $f_t$ \citep{sutton1996generalization}. Experimentally, we found $\ep=0$ optimal on average, consistent with the results of \citet{bietti:hal-01708310}.
\item[$\ep$-decreasing:] selects a random action with probabilities $\ep_i$, where $\ep_i = {\ep_0/ t}$, 
    $\ep_0 \in ]0,1]$ and $t$ is the index of the current round. In our experiments we set $\ep_0=0.1$.~\citep{Sutton:1998:IRL:551283}
\item[Exponentiated Gradient $\ep$-greedy:] maintains a set of candidate values for $\ep$-greedy exploration. At each 
    iteration, it runs a sampling procedure to select a new $\ep$ from a finite set of candidates. The probabilities
    associated with the candidates are initialized uniformly and updated with the Exponentiated Gradient (EG) algorithm.
    Following \cite{Li:2010:EEP:1835804.1835811}, we use the 
    candidate set $\{\ep_i = 0.05 \times i + 0.01, i = 1, \cdots, 10\}$ for $\ep$.
\item[LinUCB:] Maintains confidence bounds for reward payoffs and selects actions with the highest confidence bound. 
    It is impractical to run ``as is'' due to high-dimensional matrix inversions. We use diagonal approximation to the 
    covariance when the dimensions exceeds $150$.~\citep{Li:2010:CAP:1772690.1772758}
\item[$\tau$-first:] Explore uniformly on the first $\tau$ fraction of the data; after that, act greedily.%
\item[Cover:]  Maintains a uniform distribution over a fixed number of policies. The policies are used to approximate a covering distribution 
    over policies that are good for both exploration and exploitation~\citep{agarwal14taming}.
\item[Cover Non-Uniform:] similar to Cover, but reduces the level of exploration of Cover to be more competitive with the Greedy method. Cover-Nu doesn't 
    add extra exploration beyond the actions chose by the covering policies~\citep{bietti:hal-01708310}.
\end{description}

In all cases, we select the best hyperparameters for each exploration algorithm following~\cite{bietti:hal-01708310}. 
These hyperparameters are: the choice of $\ep$ in $\ep$-greedy, $\tau$ in $\tau$-first, the number of bags, and the 
tolerance $\psi$ for Cover and Cover-NU. We set $\ep=0.0$, $\tau=0.02$, $\textrm{bag size}=16$, and $\psi=0.1$.

\subsection{Experimental Results} \label{sec:comparison}

The overall results are shown in \autoref{fig:overall}. In the left-most figure, we see the CDFs for the different 
algorithms. To help read this, note that at $x=1.0$, we see that \ourname has a relative reward at least $1.0$ on 
more than 40\% of datasets, while $\ep$-decreasing and $\ep$-greedy achieve this on about 30\% of datasets.  
We find that the two strongest baselines are $\ep$-decreasing and $\ep$-greedy (better when reward differences 
are small, toward the left of the graph). The two curves for $\ep$-decreasing and $\ep$-greedy coincide. This happens
because the exploration probability $\ep_0$ for $\ep$-decreasing decays rapidly approaching zero with a rate of 
$\frac{1}{t}$, where $t$ is the index of the current round. \ourname outperforms the baselines in the ``large reward'' 
regimes (right of graph) but underperforms $\ep$-decreasing and $\ep$-greedy in low reward regimes (left of graph).
In \autoref{tab:winlose}, we show statistically-significant win/loss differences for each of the algorithms. \ourname 
is the only algorithm that always wins more than it loses against other algorithms.

To understand more directly how \ourname compares to $\ep$-decreasing, in the middle figure of \autoref{fig:overall}, we 
show a scatter plot of rewards achieved by \ourname (x-axis) and $\ep$-decreasing (y-axis) on each of the $300$ datasets, 
with statistically significant differences highlighted in red and insignificant differences in blue.
Points below the diagonal line correspond to better performance by \ourname (147 datasets) and points above to 
$\ep$-decreasing (124 datasets). The remaining 29 had no significant difference.

In the right-most graph in \autoref{fig:overall}, we show a representative example of learning curves for the various algorithms.
Here, we see that as more data becomes available, all the approaches improve (except $\tau$-first, which has ceased to learn after $2\%$ of the data).

Finally, we consider the effect that the additional features have on \ourname's performance.
In particular, we consider a version of \ourname with all features (this is the version used in all other experiments)
with an ablated version that only has access to the (calibrated) probabilities of each action from the underlying classifier $f$.
The comparison is shown as a scatter plot in \autoref{fig:ablation}.
Here, we can see that the full feature set \emph{does} provide lift over just the calibrated probabilities,
with a win-minus-loss improvement of $24$.

%% file: datasets.tex
\section{List of Datasets} \label{apx:listofdatasets}
The datasets we used can be accessed at \url{https://www.openml.org/d/<id>}. 
The list of $(\textrm{id}, \textrm{size})$ pairs below shows the (<id> for the 
datasets we used and the dataset size in number of examples:

{
\fontsize{10pt}{16pt}\selectfont
(46,100)
(716, 100)
(726, 100)
(754, 100)
(762, 100)
(768, 100)
(775, 100)
(783, 100)
(789, 100)
(808, 100)
(812, 100)
(828, 100)
(829, 100)
(850, 100)
(865, 100)
(868, 100)
(875, 100)
(876, 100)
(878, 100)
(916, 100)
(922, 100)
(932, 100)
(1473, 100)
(965, 101)
(1064, 101)
(956, 106)
(1061, 107)
(771, 108)
(736, 111)
(448, 120)
(782, 120)
(1455, 120)
(1059, 121)
(1441, 123)
(714, 125)
(867, 130)
(924, 130)
(1075, 130)
(1141, 130)
(885, 131)
(444, 132)
(921, 132)
(974, 132)
(719, 137)
(1013, 138)
(1151, 138)
(784, 140)
(1045, 145)
(1066, 145)
(1125, 146)
(902, 147)
(1006, 148)
(969, 150)
(955, 151)
(1026, 155)
(745, 159)
(756, 159)
(1085, 159)
(1054, 161)
(748, 163)
(747, 167)
(973, 178)
(463, 180)
(801, 185)
(1164, 185)
(788, 186)
(1154, 187)
(941, 189)
(1131, 193)
(753, 194)
(1012, 194)
(1155, 195)
(1488, 195)
(446, 200)
(721, 200)
(1124, 201)
(1132, 203)
(40, 208)
(733, 209)
(796, 209)
(996, 214)
(1005, 214)
(895, 222)
(1412, 226)
(820, 235)
(851, 240)
(464, 250)
(730, 250)
(732, 250)
(744, 250)
(746, 250)
(763, 250)
(769, 250)
(773, 250)
(776, 250)
(793, 250)
(794, 250)
(830, 250)
(832, 250)
(834, 250)
(863, 250)
(873, 250)
(877, 250)
(911, 250)
(918, 250)
(933, 250)
(935, 250)
(1136, 250)
(778, 252)
(1442, 253)
(1449, 253)
(1159, 259)
(450, 264)
(811, 264)
(336, 267)
(1152, 267)
(53, 270)
(1073, 274)
(1156, 275)
(880, 284)
(1121, 294)
(43, 306)
(818, 310)
(915, 315)
(1157, 321)
(1162, 322)
(925, 323)
(1140, 324)
(1144, 329)
(1011, 336)
(1147, 337)
(1133, 347)
(337, 349)
(59, 351)
(1135, 355)
(1143, 363)
(1048, 369)
(860, 380)
(1129, 384)
(1163, 386)
(900, 400)
(906, 400)
(907, 400)
(908, 400)
(909, 400)
(1025, 400)
(1071, 403)
(1123, 405)
(1160, 410)
(1126, 412)
(1122, 413)
(1127, 421)
(764, 450)
(1065, 458)
(1149, 458)
(1498, 462)
(724, 468)
(814, 468)
(1148, 468)
(1150, 470)
(765, 475)
(767, 475)
(1153, 484)
(742, 500)
(749, 500)
(750, 500)
(766, 500)
(779, 500)
(792, 500)
(805, 500)
(824, 500)
(838, 500)
(855, 500)
(869, 500)
(870, 500)
(879, 500)
(884, 500)
(886, 500)
(888, 500)
(896, 500)
(920, 500)
(926, 500)
(936, 500)
(937, 500)
(943, 500)
(987, 500)
(1470, 500)
(825, 506)
(853, 506)
(872, 506)
(717, 508)
(1063, 522)
(954, 531)
(1467, 540)
(1165, 542)
(1137, 546)
(335, 554)
(333, 556)
(947, 559)
(949, 559)
(950, 559)
(951, 559)
(826, 576)
(1004, 600)
(334, 601)
(1158, 604)
(770, 625)
(997, 625)
(1145, 630)
(1443, 661)
(774, 662)
(795, 662)
(827, 662)
(931, 662)
(292, 690)
(1451, 705)
(1464, 748)
(37, 768)
(1014, 797)
(970, 841)
(994, 846)
(841, 950)
(50, 958)
(1016, 990)
(31, 1000)
(715, 1000)
(718, 1000)
(723, 1000)
(740, 1000)
(743, 1000)
(751, 1000)
(797, 1000)
(799, 1000)
(806, 1000)
(813, 1000)
(837, 1000)
(845, 1000)
(849, 1000)
(866, 1000)
(903, 1000)
(904, 1000)
(910, 1000)
(912, 1000)
(913, 1000)
(917, 1000)
(741, 1024)
(1444, 1043)
(1453, 1077)
(1068, 1109)
(934, 1156)
(1049, 1458)
(1454, 1458)
(983, 1473)
(1128, 1545)
(1130, 1545)
(1138, 1545)
(1139, 1545)
(1142, 1545)
(1146, 1545)
(1161, 1545)
(1166, 1545)
(1050, 1563)
(991, 1728)
(962, 2000)
(971, 2000)
(978, 2000)
(995, 2000)
(1020, 2000)
(1022, 2000)
(914, 2001)
(1067, 2109)
(772, 2178)
(948, 2178)
(958, 2310)
(312, 2407)
(1487, 2534)
(737, 3107)
(953, 3190)
(3, 3196)
(1038, 3468)
(871, 3848)
(728, 4052)
(720, 4177)
(1043, 4562)
(44, 4601)
(979, 5000)
(1460, 5300)
(1489, 5404)
(1021, 5473)
(1069, 5589)
(980, 5620)
(847, 6574)
(1116, 6598)
(803, 7129)
(1496, 7400)
(725, 8192)
(735, 8192)
(752, 8192)
(761, 8192)
(807, 8192)
}

%% file: related.tex
%

The field of meta-learning is based on the idea of replacing hand-engineered learning heuristics with heuristics learned from data.
One of the most relevant settings for meta-learning to ours is active learning, in which one aims to learn a decision function to decide which examples, from a pool of unlabeled examples, should be labeled.
Past approaches to meta-learning for active learning include reinforcement learning-based strategies \citep{woodward2017active,fang2017learning}, imitation learning-based strategies \citep{bachman2017learning}, and batch supervised learning-based strategies \citep{konyushkova2017metaactive}.
Similar approaches have been used to learn heuristics for optimization \citep{li2016learning,andrychowicz2016learning},
multiarm (non-contextual) bandits \cite{maes2012meta},
and neural architecture search \citep{zoph2016neural}, recently mostly based on (deep) reinforcement learning. 
While meta-learning for contextual bandits is most similar to meta-learning for active learning, there is a fundamental difference that makes it significantly more challenging:
in active learning, the goal is to select as few examples as you can to learn, so by definition the horizon is short;
in contextual bandits, learning to explore is fundamentally a long-horizon problem, because what matters is not immediate reward but long term learning.

In reinforcement learning, \citet{gupta2018meta} investigated the task of meta-learning an exploration strategy for a distribution of related tasks by learning a latent exploration space.
Similarly, \citet{xu2018learning} proposed a teacher-student approach for learning to do exploration in off-policy reinforcement learning.
While these approaches are effective if the distribution of tasks is very similar and the state space is shared among different tasks, they fail to generalize when the tasks are different.
Our approach targets an easier problem than exploration in full reinforcement learning environments, and can generalize well across a wide range of different tasks with completely unrelated features spaces.

There has also been a substantial amount of work on constructing ``good'' exploration policies, in problems of varying complexity: traditional bandit settings \citep{NIPS2016_6341}, contextual bandits \citep{pmlr-v51-feraud16} and reinforcement learning \citep{NIPS2016_6501}.
In both bandit settings, most of this work has focused on the learning theory aspect of exploration: what exploration distributions \emph{guarantee} that learning will succeed (with high probability)?
\ourname, lacks such guarantees: in particular, if the data distribution of the observed contexts ($\phi(f_t)$) in some test problem differs substantially from that on which \ourname was trained, we can say nothing about the quality of the learned exploration.
Nevertheless, despite fairly substantial distribution mismatch (synthetic $\rightarrow$ real-world), \ourname works well in practice, and our stylized theory (\autoref{sec:guarantees}) suggests that there may be an interesting avenue for developing strong theoretical results for contextual bandit learning with learned exploration policies, and perhaps other meta-learning problems.

In conclusion, we presented \ourname, a meta-learning algorithm for learning exploration policies in the contextual bandit setting. \ourname enjoys no-regret guarantees, and empirically it outperforms alternative exploration algorithm in most settings. One limitation of \ourname is the computational resources required during the offline training phase on the synthetic datasets. In the future, we will work on improving the computational efficiency for \ourname in the offline training phase and scale the experimental analysis to problems with larger number of classes.